\documentclass[10pt,twocolumn,letterpaper]{article}

\usepackage{cvpr}
\usepackage{times}
\usepackage{epsfig}
\usepackage{graphicx}
\usepackage{amsmath}
\usepackage{amssymb}
\usepackage{authblk}

\usepackage[breaklinks=true,bookmarks=false]{hyperref}

\cvprfinalcopy 

\def\cvprPaperID{****} 

\begin{document}

\title{A Spatiotemporal Volumetric Interpolation Network for 4D Dynamic Medical Image}

\author[1]{Yuyu Guo}
\author[2]{Lei Bi}
\author[2]{Euijoon Ahn}
\author[2]{Dagan Feng}
\author[1*]{Qian Wang}
\author[2*]{Jinman Kim}
\affil[1]{Institute for Medical Imaging Technology, School of Biomedical Engineering, Shanghai Jiao Tong University, China}
\affil[2]{School of Computer Science, University of Sydney, Australia}
\affil[ ]{\small\it {\{yu.guo, wang.qian\}@sjtu.edu.cn, \space\space\space\{lei.bi, euijoon.ahn, dagan.feng, jinman.kim\}@sydney.edu.au}}

\maketitle

\begin{abstract}
Dynamic medical imaging is usually limited in application due to the large radiation doses and longer image scanning and reconstruction times. Existing methods attempt to reduce the dynamic sequence by interpolating the volumes between the acquired image volumes. However, these methods are limited to either 2D images and/or are unable to support large variations in the motion between the image volume sequences. In this paper, we present a spatiotemporal volumetric interpolation network (SVIN) designed for 4D dynamic medical images. SVIN introduces dual networks: first is the spatiotemporal motion network that leverages the 3D convolutional neural network (CNN) for unsupervised parametric volumetric registration to derive spatiotemporal motion field from two-image volumes; the second is the sequential volumetric interpolation network, which uses the derived motion field to interpolate image volumes, together with a new regression-based module to characterize the periodic motion cycles in functional organ structures. We also introduce an adaptive multi-scale architecture to capture the volumetric large anatomy motions. Experimental results demonstrated that our SVIN outperformed state-of-the-art temporal medical interpolation methods and natural video interpolation methods that has been extended to support volumetric images. Our ablation study further exemplified that our motion network was able to better represent the large functional motion compared with the state-of-the-art unsupervised medical registration methods.
\end{abstract}

\section{Introduction}

Dynamic medical imaging modalities enable the examination of functional and mechanical properties of the human body and are used for clinical applications, e.g., four-dimensional (4D) computed tomography (CT) for respiratory organ motion modelling \cite{pan20044d}, 4D magnetic resonance (MR) imaging for functional heart analysis \cite{bornstedt2001multi}, and 4D ultrasound (US) for echocardiography analysis \cite{zhang2011spatio}. These 4D modalities have high spatial (volumetric) and temporal (time sequence) sampling rate to capture the periodic motion cycles of organ activities, and this information is used for clinical decision making. However, the acquisition of these dynamic images requires larger radiation doses which may cause harm to humans, and longer image scanning and reconstruction times; these factors limit the use of 4D imaging modalities to broader clinical applications \cite{ohno2002solitary,cane20184d}. 

\begin{figure}[h]
\begin{center}
\includegraphics[width=0.9\linewidth]{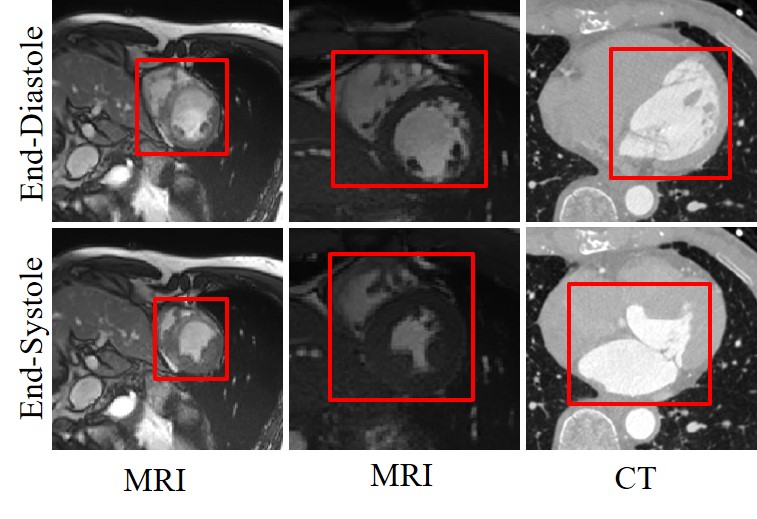}
\end{center}
   \caption{The cardiac motions in two-time phases: End-Systole (ES) and End-Diastole (ED). The red bounding boxes highlight the heart structure. All images showing transaxial views, cropped to enlarge the heart.}
\label{fig:1}
\end{figure}

To mitigate these factors, reducing the temporal sampling has been widely employed but this compromises valuable temporal information \cite{li2008advances,hollingsworth2015reducing}. In these approaches, intermediary image volumes are interpolated from their adjacent volumes. Such interpolation methods are reliant on either non-rigid registration \cite{baumgartner2013groupwise,metz2011nonrigid,zhang2011spatio} or optical flow-based \cite{nam2006optical,karani2017temporal} algorithms. Non-rigid registration approaches calculate the dense image volume correspondences that occur from one volume to another, and then uses the calculated correspondences to generate the intermediary volumes. Such approaches, however, often generate artifacts or fuzzy boundaries and do not perform well when the variations in anatomy or organ activity (e.g., size and shape) are large. An alternative approach was to use optical flow-based methods (using deep learning) \cite{karani2017temporal,yan2018left} to estimate a dense motion (i.e., deformation) field between image pairs. However, these methods were limited to 2D image interpolation and therefore did not utilize the rich spatial information inherent in medical image volumes. They are also limited when the motion between the image sequences are not in linear trajectory and are not changing in a constant velocity. Therefore, these approaches are not applicable to volumetric temporal imaging modalities that exhibit large non-linear motions in spatiotemporal space.


In this paper we propose a spatiotemporal volumetric interpolation network (SVIN) designed for 4D dynamic medical images. To the best of our knowledge, this is the first deep learning-based method for 4D dynamic medical image interpolation. An overview of our model is illustrated in Fig. \ref{fig:2} which comprises of two main networks. Our first spatiotemporal motion network leverages the 3D convolutional neural network (CNN) for unsupervised parametric volumetric registration to derive spatiotemporal motion field from two-image volumes. In the second sequential volumetric interpolation network, the derived motion field is used to interpolate the image volume, together with  a new regression-based module to characterize the periodic motion cycles in functional organ structures. We also propose an adaptive multi-scale architecture that learns the spatial and appearance deformation in multiple volumes to capture large motion characteristics. We demonstrate the application of our method on cardiac motion interpolation, which is acquired using both 4D CT and 4D MR images. These images are characterized by twisting action during contraction to relaxation of the heart structure, and has complex changes in muscle morphology, as depicted in Fig. \ref{fig:1}. Our method was used to increase the temporal resolution in both the CT and MR image volumes. We evaluate our method in comparison to the state-of-the-art interpolation method. We further conducted an ablation study to demonstrate the effectiveness of our motion network.

\section{Related Works}

We partitioned the related works into three categories which we deemed relevant to our research: (1) Medical dynamic image interpolation; (2) spatiotemporal motion field calculation for medical image and (3) natural video interpolation approaches.

\subsection{Dynamic medical image interpolation}

Many existing medical image interpolation methods rely upon optical flow-based or non-rigid registration methods to generate a linearly interpolated image by averaging pixel values between the adjacent image sequences \cite{baumgartner2013groupwise,lee2003real,nam2006optical,metz2011nonrigid,zhang2011spatio,tryggestad2013respiration}. For instance, Ehrhardt et al. \cite{ehrhardt2006optical} presented an optical flow-based method to establish spatial correspondence between adjacent slices for cardiac temporal image. Zhang et al. \cite{zhang2011spatio} used non-rigid registration-based method to synthesize echocardiography and cardiovascular MR image sequences. The main advantage of these approaches is that they track spatiotemporal motion field, in a pixel-wise manner, between the neighboring images to estimate the interpolation. However, their assumption limited the spatiotemporal motion between the adjacent images to be in a linear trajectory, and thus disregarded the complex, non-linear motions apparent in functional organ structures. Recently, there are two CNN based methods for temporal interpolation via motion field for MR images from Lin Zhang et al. \cite{lin2018temporal} and Kim et al. \cite{karani2017temporal}. They achieved outstanding performance compared with previous works. However, their method did not support full 3D volumetric information and did not perform well when there was large variations in motion. 

\subsection{Learning spatiotemporal motion fields from volume image sequence}

Many studies used deformable medical image registration techniques to estimate the motion field between the input image sequences. The deformable medical image registration techniques can be divided into two parts: non-learning based \cite{thirion1998image,ashburner2007fast,klein2009evaluation,balci2007free,dryden2014shape} and learning-based methods \cite{krebs2017robust,yang2017quicksilver,sokooti2017nonrigid}. The typical non-learning based approaches are free-form deformations with B-splines \cite{ashburner2007fast}, Demons \cite{thirion1998image} and ANTs \cite{avants2011reproducible}. These approaches optimize displacement vector fields by calculating the similarity of the topological structures. Deep learning-based methods, in recent years, used labelled data of spatiotemporal motion field and have shown great performances \cite{krebs2017robust,yang2017quicksilver,sokooti2017nonrigid}. However, their performance was dependent upon the availability of large-scale labelled data. To address this, several unsupervised methods were proposed to predict the spatiotemporal motion field \cite{de2017end,li2017non,balakrishnan2018unsupervised}. Although these methods demonstrated promising results, \cite{de2017end} and \cite{li2017non} were only useful in  patch-based volumes or in 2D slices. Jiang et al. \cite{balakrishnan2018unsupervised} recently developed a CNN, VoxelMorph which used full 3D volumetric information. However, it was not designed for dynamic image sequences where it has large variations in motion.

\begin{figure*}[h]
\begin{center}
\includegraphics[width=0.9\linewidth]{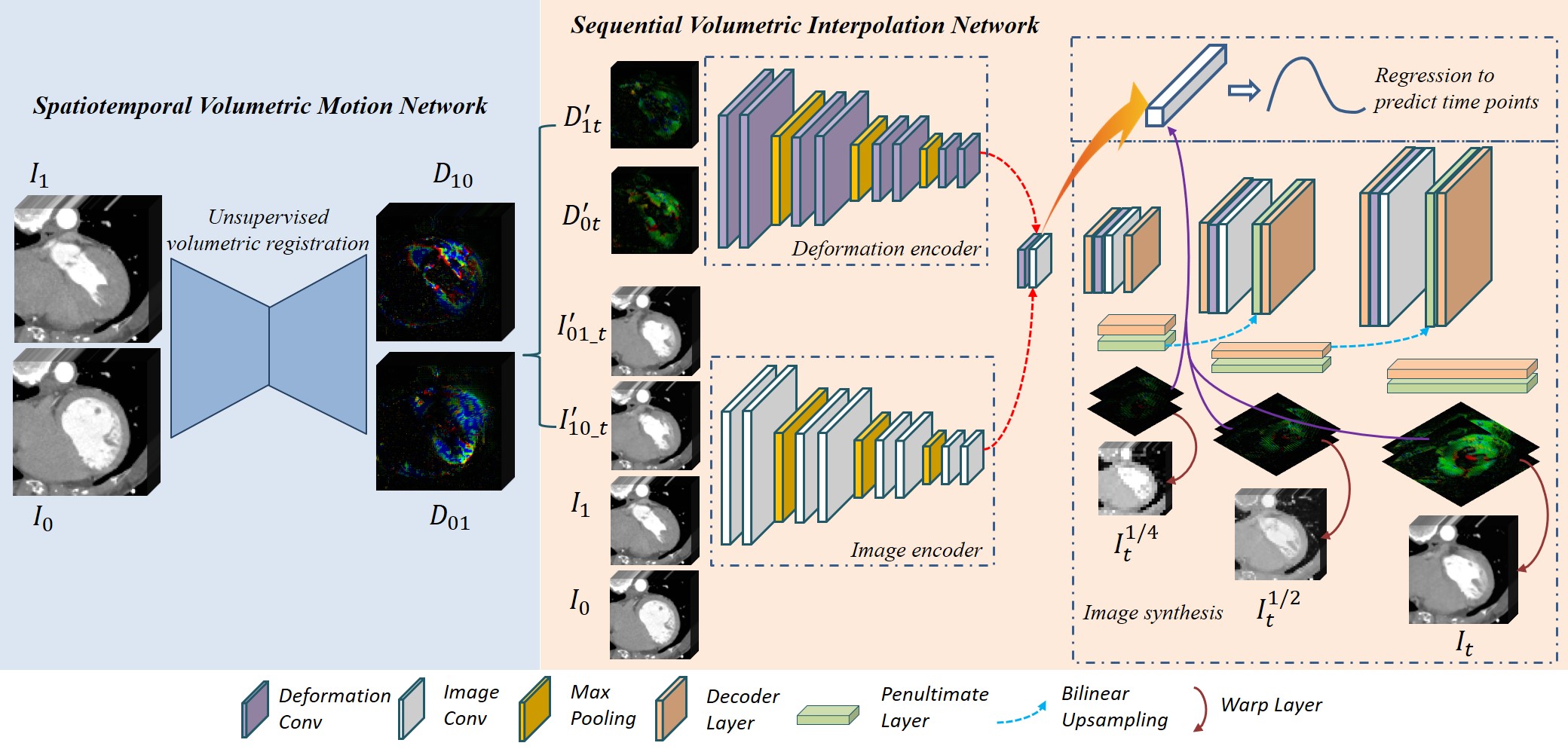}
\end{center}
   \caption{An overview of the proposed method which contains a motion network and an interpolation network. An adaptive multi-scale architecure is used in both of motion and interpolation network to cover the large motion. A regression module is intergrated in our interpolation network to constrain the intermediated motion field.}
\label{fig:2}
\end{figure*}

\subsection{Natural video interpolation approaches}
Video interpolation is an active research task in natural scenes, e.g., model-based tracking, patch identification, and matching and framerate upsampling \cite{jeon2003coarse,choi2007motion,long2016learning,meyer2018phasenet}. Niklaus et al. \cite{niklaus2017video} developed a spatially-adaptive convolution kernel to estimate the motion for each pixels. Liu et al. \cite{liu2017video} divided the frame interpolation into two steps, optical flow estimation and image interpolation. Their network learnt an input pair of consecutive frames in an unsupervised manner and then refined the interpolation based on the outputs of the estimation. Jiang et al. \cite{jiang2018super} presented Slomo – a technique which interpolates frame motion by linearly combining bi-directional optical flows, and then further refining the estimated motion flow field through an end-to-end CNN. Recently, Peleg et al. \cite{peleg2019net} presented a multi-scale structured architecture neural network to better capture the local details from high resolutions frame. However, when considering the application of these methods to dynamic medical images interpolation, this is a challenging problem as the temporal sampling in medical image volume sequences are much lower than that of natural scene videos. In addition, the deformation and visual differences in dynamic medical images are comparatively more complex and non-trivial than natural scene videos.

\section{Proposed Method}


Let $\left\{\textit{$I_{T}$, $T=1,2...,N$}\right\}$ be a sequence of volumetric images representing the cardiac motion from end-diastole (ED) (\textit{$T=1$}) to end-systole (ES) (\textit{$T=N$}) phase, and let $\left\{\textit{$I_{i}$, $I_{j}$ $|$ ($i$, $j$) $\in$ $T$}\right\}$ be a pair of cardiac images indicating two random time points within the cardiac motion. Our aim is to interpolate the intermediate image \textit{$I_{t}, (t \in T)$}. For this work, we used images at ED (denote as \textit{$I_{ED}$}) and ES (denote as \textit{$I_{ES}$}) phase to interpolate the complete the cardiac motion. $\left\{\textit{$\phi_{i \rightarrow j}$, $\phi_{j \rightarrow i}$}\right\}$ denotes the motion field between \textit{$I_{i}$} and \textit{$I_{j}$} in bi-directions. 

\begin{figure*}[htbp]
\begin{center}
\includegraphics[width=0.8\linewidth]{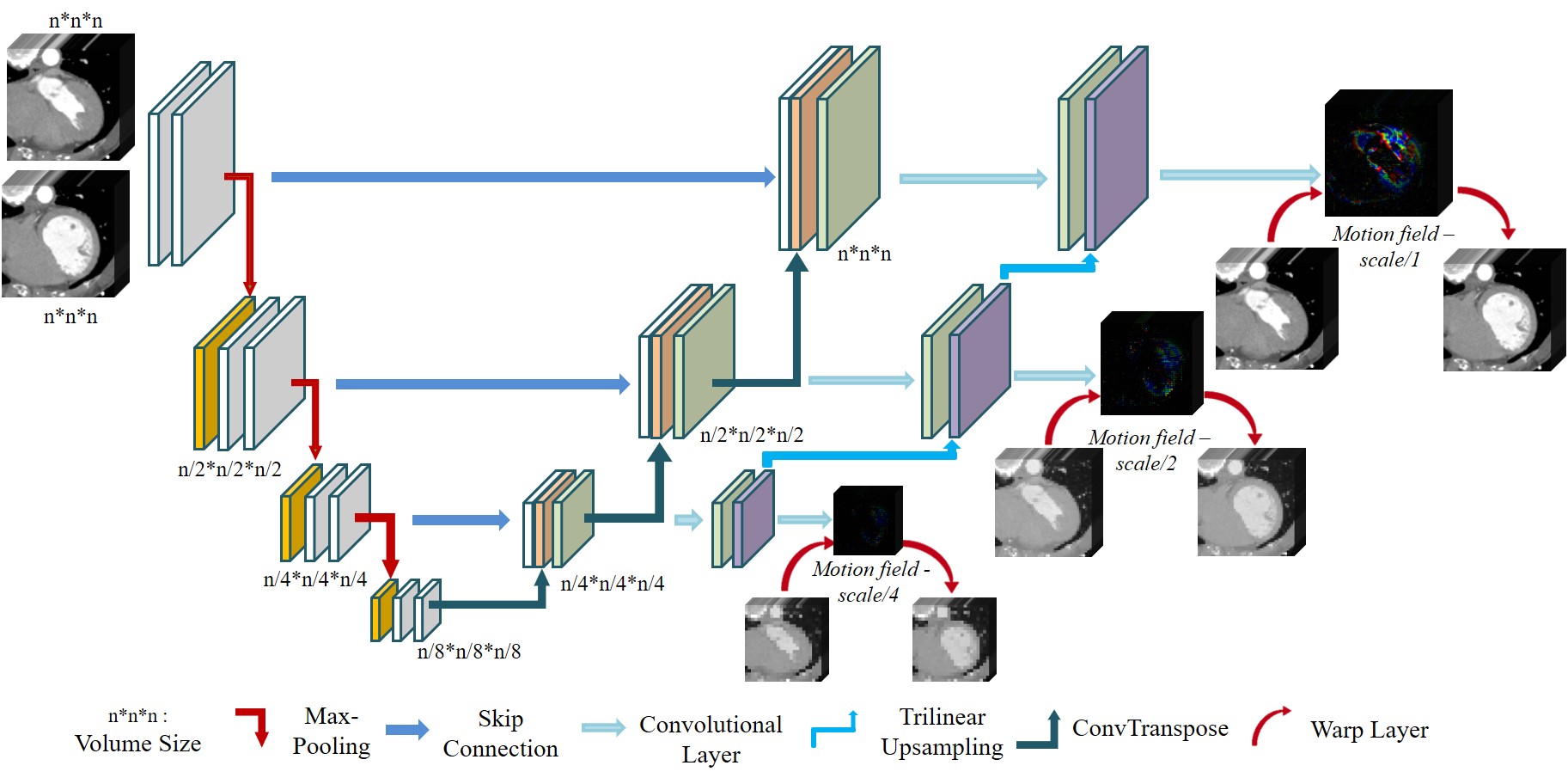}
\end{center}
   \caption{The architecture of our spatiotemporal volumetric motion network with an adaptive multi-scale architecture.}
\label{fig:3}
\end{figure*}

Fig. \ref{fig:2} shows the overall proposed method. Initially, spatiotemporal motion network was used to learn and capture bi-directional motion fields between \textit{$I_{ED}$} and \textit{$I_{ES}$} in an unsupervised manner. Two linearly interpolated intermediate images were then coarsely created using the learned spatiotemporal motion fields, \textit{$\phi_{ED \rightarrow ES}$} and \textit{$\phi_{ES \rightarrow ED}$}. Using the coarsely interpolated intermediate images and their corresponding deformation fields, we further refined the coarse intermediate images by the volumetric interpolation network, where we used a regression-based module to constrain the interpolation to follow the patterns of cardiac biological motion. Specifically, both of our volumetric motion estimation and interpolation network are using an adaptive multi-scale architecture which enables to capture various types motions - both small and large volume spatiotemporal deformations (see Fig. \ref{fig:2} and \ref{fig:3}).


\subsection{Spatiotemporal volumetric motion field estimation}


Fig. \ref{fig:3} presents the architecture of 3D CNN for spatiotempopral motion field estimation. We estimate a motion field that can represent the voxel-wise motion flow of volume images at two individual time points. This can be represented as a function \textit{$D_{\theta}(I_{i},I_{j})=\phi_{i \leftrightarrow j}(\Delta x,\Delta y,\Delta z)$}, where \textit{$\phi_{i \leftrightarrow j}(\Delta x,\Delta y,\Delta z)$} indicates the vectors that represent the movement in 3D space and \textit{$\theta$} are the learnable parameters of the network. 
We used an encoder-decoder architecture with skip connections for generating \textit{$\phi_{i \leftrightarrow j}$} by given \textit{$I_{i}$} and \textit{$I_{j}$}.

In order to produce a volumetric motion field that can cover various types of deformations, we propose an adaptive multi-scale architecture that embeded both a global and a local learning. More specifically, for global learning, our motion field estimation network focuses on large deformation while the volumetric images in a low-scale level would ignore the local details while more detailed information will be covered when the volumetric image in a high-scale. In addition, the global deformation from low-scale is integrated to high-scale, which reduces the difficulty of the network for learning and constrains the network to pay more attention to detailed deformation. 
Our deformation field can be defined as:

\begin{equation}\label{eq:1}
\begin{aligned}
    \phi_{i \rightarrow j}&=D_{\theta}(I_{i},I_{j}) \quad\text{or}\quad \phi_{j \rightarrow i}&=D_{\theta}(I_{j},I_{i})
\end{aligned}
\end{equation}

\begin{equation}\label{eq:2}
\begin{aligned}
    I_{j}=\zeta(I_{i}|\phi_{i \rightarrow j})\quad\text{or}\quad I_{i}=\zeta(I_{j}|\phi_{j \rightarrow i})
\end{aligned}
\end{equation}

where \textit{$\zeta(I_{i})|\phi_{i \rightarrow j}$} representing the warped image by the spatial vector field \textit{$\phi_{i \rightarrow j}$} with bilinear interpolation. 

For training our motion field estimation network, we used an image-wise similarity loss and a motion field smoothness regularization loss with an adaptive multi-scale network architecture (as shown in Fig. \ref{fig:3}). Given the network output \textit{$\phi_{i \rightarrow j}^{n}\left\{n=1,2,3\right\}$}, where \textit{$i$} denotes the volumetric images at different scales (we used 3 different scales in total), we define a motion field smoothness regularization loss as:

\begin{equation}\label{eq:3}
\begin{aligned}
    \mathcal{L}_{\phi}(D_{\theta}(I_{i},I_{j}))=\sum_{c=1}^{3} \parallel \nabla \phi_{i \rightarrow j}^{c}\parallel_{1}
\end{aligned}
\end{equation}

 Where $\nabla$ is the gradient operator. The image-wise similarity loss was leveraged from VoxelMorph \cite{balakrishnan2018unsupervised} and this can be defined as:
 
 \begin{equation}\label{eq:4}
\begin{aligned}
    \mathcal{L}_{s}(\zeta(I_{i}|\phi_{i \rightarrow j}),I_{j})=\sum_{c=1}^{3} \parallel \zeta(I_{i}^{c}|\phi_{i \rightarrow j}^{c})-I_{j}^{c}\parallel_{2}
\end{aligned}
\end{equation}

 \subsection{Sequential volumetric interpolation network}
 
 
Based on the derived deformation fields \textit{$\phi_{ED \rightarrow ES}$} and \textit{$\phi_{ES \rightarrow ED}$}, we used used linear interpolation approach to synthesize the intermediate deformation fields, as shown in follows:
 
\begin{equation}\label{eq:5}
\begin{aligned}
    \phi_{ED \rightarrow t}=t\phi_{ED \rightarrow ES}
\end{aligned}
\end{equation}
\begin{equation}\label{eq:6}
\begin{aligned}
    \phi_{ES \rightarrow t}=(1-t)\phi_{ES \rightarrow ED}
\end{aligned}
\end{equation}
 
 Based on Eqs. \ref{eq:3} and \ref{eq:4}, the linear interpolation based deformation field for \textit{$I_{t}$} can be approximated as:
 
\begin{equation}\label{eq:7}
\begin{aligned}
    \tilde{I_{t}}=(1-t)\zeta(I_{ED}|\phi_{ED \rightarrow t})+t\zeta(I_{ES}|\phi_{ES \rightarrow t})
\end{aligned}
\end{equation}
 
 To improve the consistency in bi-directions, Eq. \ref{eq:3} and \ref{eq:4} can be modified to as follows:
 
\begin{equation}\label{eq:8}
\begin{aligned}
    \phi_{ED \rightarrow t}=&t(1-t)\phi_{ED \rightarrow ES}\\
    &-t^{2}\zeta(\phi_{ES \rightarrow ED}|\phi_{ES \rightarrow ED})
\end{aligned}
\end{equation}
\begin{equation}\label{eq:9}
\begin{aligned}
    \phi_{ES \rightarrow t}=&-(1-t)^{2}\zeta(\phi_{ED \rightarrow ES}|\phi_{ED \rightarrow ES})\\
    &+t(1-t)\phi_{ES \rightarrow ED}
\end{aligned}
\end{equation}
 
In addition, we introduce a hyper-weight map \textit{$\gamma$} to balance the importance of using deformation from the bi-directions (forward and backward directions) and this can be defined as:
 
\begin{equation}\label{eq:10}
\begin{aligned}
    \gamma_{ES}=1-\gamma_{ED}
\end{aligned}
\end{equation}
 
 Thus, the linear image interpolation based on bi-directional deformation and \textit{$\gamma$} can be defined as:
 
\begin{equation}\label{eq:11}
\begin{aligned}
    \tilde{I_{t}}=&(1-t)\gamma_{ED}\zeta(I_{ED}|\phi_{ED \rightarrow t})\\
    &+t\gamma_{ES}\zeta(I_{ES}|\phi_{ES \rightarrow t})
\end{aligned}
\end{equation}
 
 As examplified in right-side of Fig. \ref{fig:2}, we used an adaptive multi-scale network architecture to ensure the synthesized intermediate volumetric images will have a high spatial-temporal resolution.

 \subsection{Regression-based module for interpolation constraints}
\begin{figure}[!htbp]
\begin{center}
\includegraphics[width=0.7\linewidth]{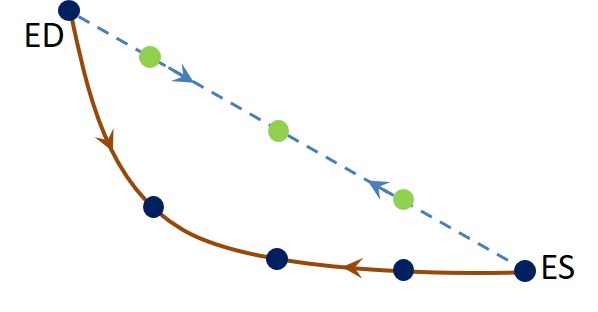}
\end{center}
   \caption{Illustration of left ventricle (LV) volume changing during the cardiac contraction period. The brown curve shows the real motion flow of LV, and blue hidden line shows the simple linear assumption. The blue points and green points represent the intermediate time points.}
\label{fig:4}
\end{figure}

 Since most biological movements have a relatively fixed motion pattern, especially in cardiac motion \cite{biffi2017investigating}, we present a regression-based module to model the relationship between cardiac motion of the cardiac cycle and time phase (as shown in Fig. \ref{fig:4}). 
 Specifically, we attempted to build a regression model representing the population-based cardiac  motion vector which indicate the shape variability at individual time point. The population-based cardiac motions at individual time point was then used to constrain the appearance of the synthetic intermediate volumetric images. Our regression estimation \textit{$R_{\theta}$} at time point \textit{$\tilde{t}$} is defined as:

\begin{equation}\label{eq:12}
\begin{aligned}
    \tilde{t} = R_{\theta}(&\phi_{ED \rightarrow t}-t\phi_{ED \rightarrow ES},\\ &\phi_{ES \rightarrow t}-(1-t)\phi_{ES \rightarrow ED})
\end{aligned}
\end{equation}
 
\subsection{Training details for volumetric interpolation}

For training our sequential volumetric interpolation network, our loss function is defined as a sum of an image-wise similarity loss \textit{$L_{similar}$}, a regression loss  \textit{$L_{r}$} and a regulation loss \textit{$L_{g}$}:

\begin{equation}\label{eq:13}
\begin{aligned}
    \mathcal{L}=\lambda_{s}\mathcal{L}_{similar}+\lambda_{r}\mathcal{L}_{r}+\lambda_{g}\mathcal{L}_{g}
\end{aligned}
\end{equation}

where image-wise similarity loss \textit{$L_{similar}$} is used to evaluate the similarity of the predicted synthetic intermediate images and the real intermediate images at multiple image scales  and is defined as: 

\begin{equation}\label{eq:14}
\begin{aligned}
    \mathcal{L}_{similar}=\sum_{c=1}^{3}\sum^{N}_{k=1} \parallel \tilde{I}^{c}_{t_{k}}-I^{c}_{t_{k}}\parallel_{2}
\end{aligned}
\end{equation}

where \textit{$\sum_{c=1}^{3}()$} represents a 3-scales volumetric image loss. \textit{$\left\{I_{t_{n}}\right\}^{N}_{n=1}$} represents the real intermediate volumetric images and \textit{$\left\{\tilde{I}_{t_{n}}\right\}^{N}_{n=1}$} represents the predicted synthetic intermediate volumetric images. The regression loss \textit{$L_{r}$} is defined as the appearance difference at individual time point:

\begin{equation}\label{eq:15}
\begin{aligned}
    \mathcal{L}_{r}=\sum^{N}_{k=1} \parallel \tilde{t}_{k}-t_{k}\parallel_{1}
\end{aligned}
\end{equation}

Regularization loss \textit{$L_{g}$}  is used to constrain the predicted motions to be consistent in bi-directions and is defined as:

\begin{equation}\label{eq:16}
\begin{aligned}
    \mathcal{L}_{g}=\sum^{3}_{c=1} \parallel \nabla\phi_{ED \rightarrow t}^{c}+\nabla\phi_{ES \rightarrow t}^{c}\parallel_{1}
\end{aligned}
\end{equation}

The weights \textit{$\lambda_{r}=1;\lambda_{s}=500;\lambda_{g}=50$} have been set empirically using a validation set. 

\section{Experiments}

\subsection{Materials and implementation details}

\begin{figure}[!htbp]
\begin{center}
\includegraphics[width=0.9\linewidth]{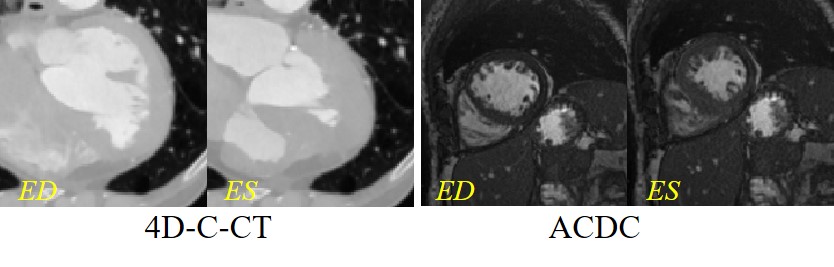}
\end{center}
   \caption{A snapshot of our training data}
\label{fig:5}
\end{figure}

We demonstrate our method with two datasets: 4D Cardiac CT (4D-C-CT), and ACDC (4D-MR cardiac cine or tagged MR imaging) \cite{bernard2018deep}.  Fig. \ref{fig:5} shows a snapshot of randomly sampled cardiac sequence volume slices. The 4D-C-CT dataset consists of 18 patient data, each having 5 time points (image volumes) from ED to ES. Image volume is characterized by a high-resolution ranging from 0.32 to 0.45 mm in intra-slice (x- and y-resolutions) and from 0.37mm to 0.82mm in inter-slice (z-resolution). The ACDC dataset contains 100 patients data. On average, each patient has 10.93 time points from ED to ES and it has an imaging resolution from 1.37 to 1.68 mm in x- and y-resolution and 5 to 10 mm in z-resolution. All scans of 4D-C-CT were resampled to a 128x128x96 grid and crop resulting images to 96x96x96. For ACDC dataset, we resampled all scans to 160x160x10. We pad ACDC data in z-axis by 0, increasing its size to 160x160x12 to reduce the border effects of 3D convolution. We randomly selected 80 training / 20 testing patient data and applied contrast-normalization to both of datasets, consistent to other similar researches \cite{jang2017automatic}.

We implemented all the networks using Pytorch library and was trained on two 11GB Nvidia 1080Ti GPUs. All models were trained with a learning rate of 0.0001. In all our evaluations, we used 3-fold cross-validation on both the datasets.


\subsection{Evaluation and metrics}

In order to evaluate the two networks in our SVIN, we conducted an ablation study. For the unsupervised spatiotemporal motion network, we compared it with state-of-the-art CNN based deformable medical image registration – VoxelMorph \cite{balakrishnan2018unsupervised}. For the interpolation network, state-of-the-art image interpolation methods were used in the comparison including (i) RVLI \cite{zhang2011spatio} – registration-based volume linear interpolation for medical images, (ii) MFIN \cite{lin2018temporal} – CNN-based medical image interpolation (2D slice-based), and (iii) Slomo – natural video interpolation \cite{jiang2018super} in 2D as well as its extension to work on medical image volumes (3D-Slomo). For image volume interpolation, we interpolated 3 intermediate volumes in between the ED-ES frames (see Fig \ref{fig:5}), evenly distributed across the time points.


We used the standard image interpolation evaluation metrics including Peak Signal-to-Noise Ratio (PSNR), Structural Similarity Index (SSIM), Mean Squared Error, and Normalized Root Mean Square Error (NRMSE). We used the same evaluation metrics for the spatiotemporal motion field estimation, consistent to other medical image registration approaches \cite{lin2018temporal}. In addition, we further used Dice Similarity Coefficient (DSC) to measure the usefulness of our interpolation in medical imaging applications. 


\section{Results and Discussion}

\begin{figure}[!htbp]
\begin{center}
\includegraphics[width=1\linewidth]{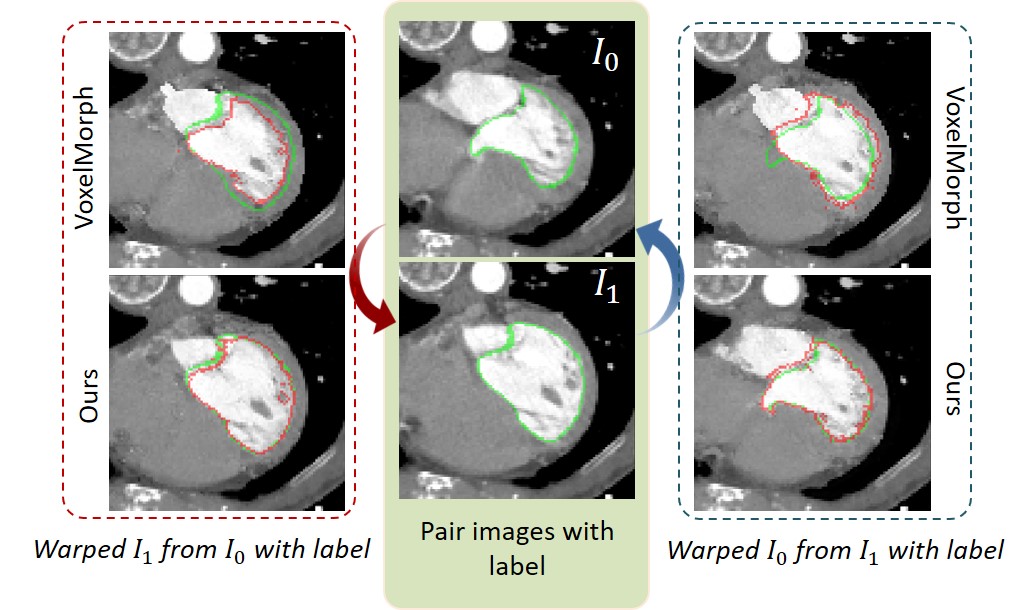}
\end{center}
   \caption{Comparison of spatiotemporal volumetric motion estimation results. The intensity image is warped from estimated spatiotemporal motion field. The red curve represents the real segmentation results while the green color shows the warped segmentation results (see the yellow arrows indicated). The red arrows indicate some organ boundaries.}
\label{fig:6}
\end{figure}

\subsection{Ablation study $–$ spatiotemporal volumetric motion field estimation}

\begin{table}[!htbp]
\caption{The performance of spatiotemporal motion field estimation on 4D-C-CT dataset.}

\begin{center}
\footnotesize
\begin{tabular}{*6c}
\hline
{} &  MSE(\textit{$10^{-2}$}) & PSNR & NRMSE & SSIM & Dsc\\
\hline
VoxelMorph & 0.787 &  27.10 & 0.276  & 0.807 & 0.880\\ 
Ours &  \textbf{0.197} & \textbf{33.17} & \textbf{0.138}  & \textbf{0.918} & \textbf{0.944}\\
\hline
\end{tabular}
\end{center}
\label{table:1}
\end{table}

\begin{table}[!htbp]
\caption{The performance of spatiotemporal motion field estimation on ACDC dataset.}
\begin{center}
\footnotesize
\begin{tabular}{*6c}
\hline
{} &  MSE(\textit{$10^{-1}$}) & PSNR & NRMSE & SSIM & Dsc\\
\hline
VoxelMorph & 0.194 &  38.06 & 0.132  & 0.912 & 0.920\\ 
Ours &  \textbf{0.168} &  \textbf{38.93} & \textbf{0.121}  & \textbf{0.914} & \textbf{0.936}\\
\hline
\end{tabular}
\end{center}
\label{table:2}
\end{table}

\begin{figure*}[!htbp]
\begin{center}
\includegraphics[width=0.9\linewidth]{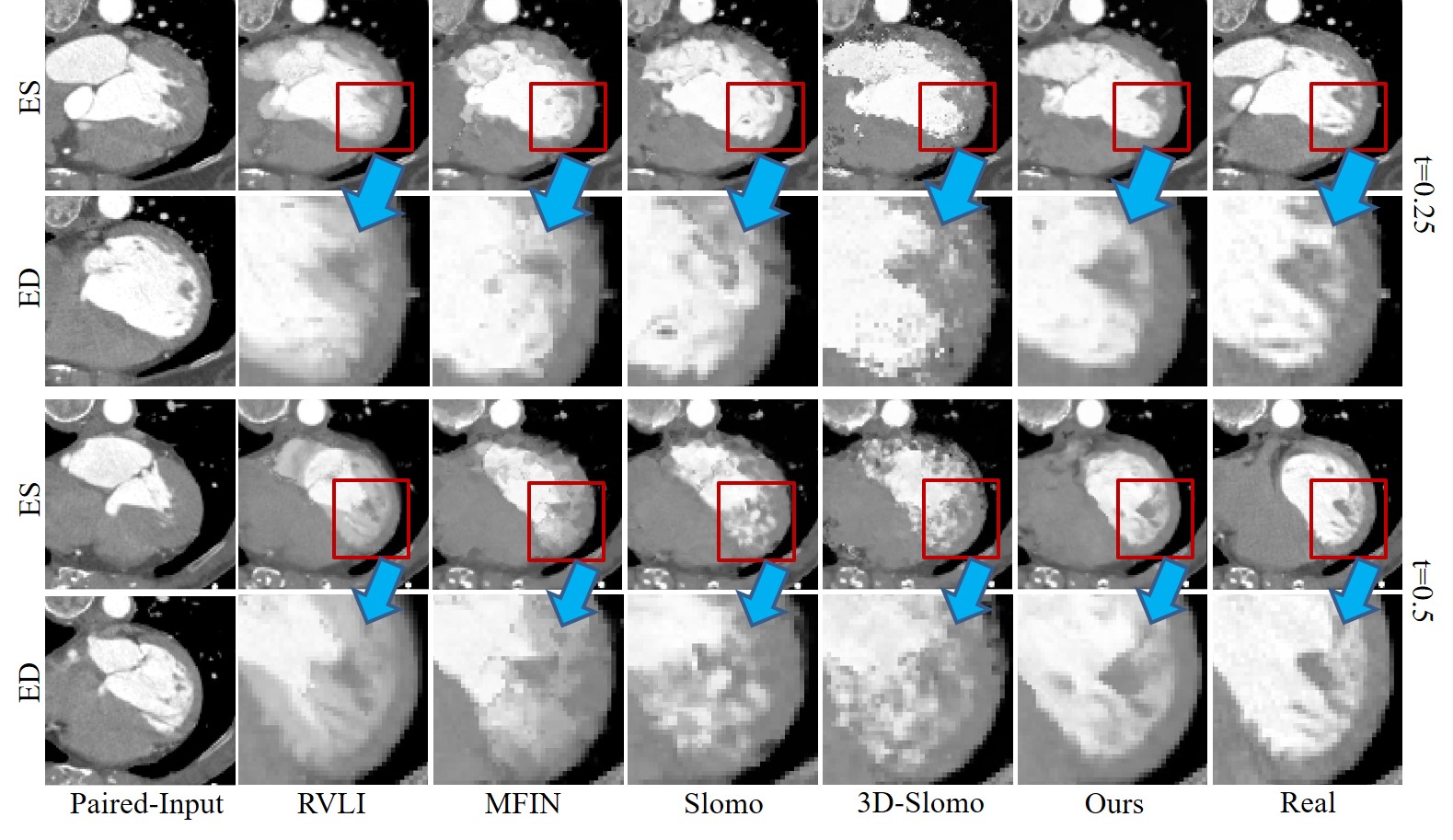}
\end{center}
   \caption{Visual results of two samples from 4D-C-CT. The first left column shows the paired-input volumes (ED and ES) and the last right column shows the real intermediate volume. The rest columns show the interpolated intermediary volumes of different approaches.}
\label{fig:7}
\end{figure*}

The results of motion field estimation on two datasets - 4D-C-CT and ACDC are shown in Table \ref{table:1} and \ref{table:2}. Our results show that motion estimation network with our adaptive architecture outperforms the recent VoxelMorph \cite{balakrishnan2018unsupervised} across all metrics on 4D-C-CT dataset, achieving the PSNR score of 33.176, NRMSE of 0.1388, SSIM of 0.9185 and MSE of 0.00197. Similarly, it also had better scores across all metrics on ACDC dataset. Our motion estimation architecture had higher improvements on 4D-C-CT dataset than that of ACDC dataset relative to VoxelMorph. We attribute this to our robust multi-scale adaptive 3D CNN which can effectively learn both large  and small variations in motion.


Fig. \ref{fig:6} shows the synthesized volumes based on the derived motion field and their corresponding warped segmentation results. It clearly shows that the warped segmentation results from the motion field learnt by our motion architecture is more similar to the ground truth.

\subsection{Comparison with the state-of-the-art interpolation methods}

Table \ref{table:3} and \ref{table:4} represent the interpolation results of different time points from ED to ES on 4D-C-CT and ACDC datasets, respectively. As expected, results show that the intermediate volumes that are in later time points had better performances. This is due to the fact that the earlier time points have larger motion variations, which contributed to its lower accuracy.


\begin{table}[!htbp]
\caption{Multi-volume cardiac sequence interpolation results on the 4D-C-CT dataset.}
\begin{center}
\small
\begin{tabular}{*5c}
\hline
{} &  MSE(\textit{$10^{-2}$}) & PSNR & NRMSE & SSIM\\
\hline
1st-point & 0.45 &  29.45 & 0.211  & 0.830\\ 
2nd-point &  0.43 &  29.47 & 0.210  & 0.825\\
3rd-point &  0.28 &  31.52 & 0.165  & 0.863\\
\hline
\end{tabular}
\end{center}
\label{table:3}
\end{table}

\begin{table}[!htbp]
\caption{Multi-volume cardiac sequence interpolation results on the ACDC dataset.}
\begin{center}
\small
\begin{tabular}{*5c}
\hline
{} &  MSE(\textit{$10^{-2}$}) & PSNR & NRMSE & SSIM\\
\hline
1st-point & 1.22 &  39.34 & 0.109  & 0.934\\ 
2nd-point &  0.95 &  40.42 & 0.087  & 0.950\\
3rd-point &  0.28 &  45.86 & 0.052  & 0.977\\
\hline
\end{tabular}
\end{center}
\label{table:4}
\end{table}

\begin{figure*}[!htbp]
\begin{center}
\includegraphics[width=0.9\linewidth]{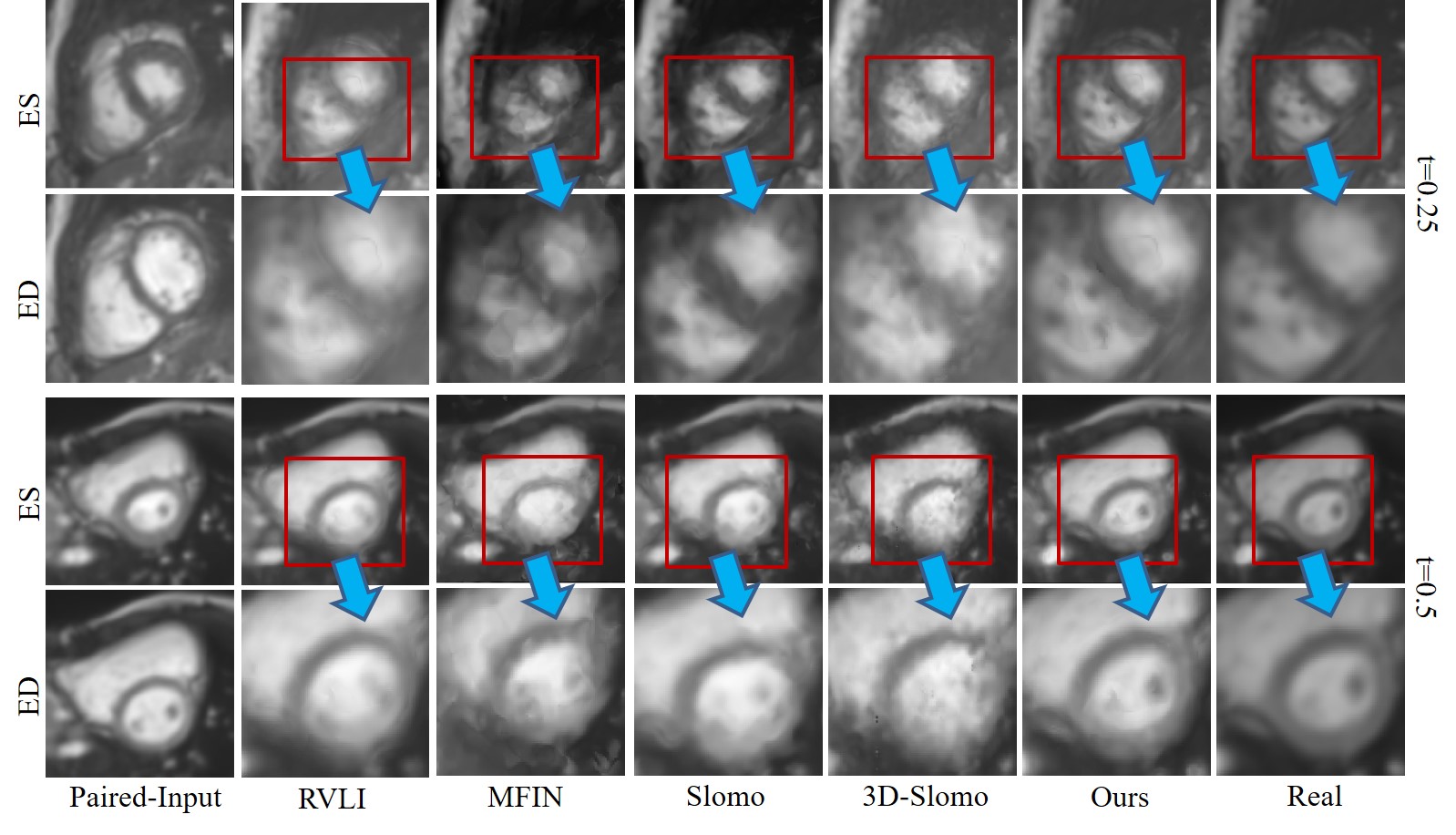}
\end{center}
   \caption{Visual results of two samples from ACDC. The first left column shows the paired-input volumes (ED and ES) and the last right column shows the real intermediate volume. The rest columns show the interpolated intermediary volumes of different approaches.}
\label{fig:8}
\end{figure*}

The comparative quantitative results for volume interpolation are shown in Table \ref{table:5} and \ref{table:6}. SVIN outperformed all other state-of-the-art interpolation method on 4D-C-CT dataset across all measures. Similarly, it also had the best scores across all metrics on the ACDC dataset. We attribute this to our adaptive multi-scale architecture capturing the variant type of motions and regression-based module which effectively constrains the intermediate volumetric motions and learn relevant inherent functional motion patterns (see Fig. \ref{fig:7} and \ref{fig:8}). Our results show that the RVLI was the closest to our results. However, the RVLI was not able to accurately interpolate the volumes when there were artifacts as evident in Fig. \ref{fig:7} and \ref{fig:8}. MFIN and Slomo also did not consider full 3D volumetric information, i.e., limited to 2D space, which contributed to its lower scores. As expected, our implemented 3D-Slomo produced a better result relative to the 2D methods. The 3D-Slomo, however, was not able to accurately synthesize the clear organ boundary and estimate the motion trajectory when there are large changes of cardiac activities (see Fig. \ref{fig:7}).

\begin{table}[!htbp]
\caption{Performance comparisons on the 4D-C-CT dataset.}
\vspace{2mm}
\begin{center}
\footnotesize
\begin{tabular}{*6c}
\hline
{} &  MSE(\textit{$10^{-2}$}) & PSNR & NRMSE & SSIM & Dsc\\
\hline
MFIN & 1.06 & 26.84 & 0.308  & 0.709 & 0.844\\ 
Slomo & 1.13 & 26.52 & 0.308  & 0.704 & 0.839\\ 
3D-Slomo & 0.92 & 26.33 & 0.303  & 0.713 & 0.872\\ 
RVLI &  0.54 &  28.70 & 0.237  & 0.806 & -\\
Ours &  \textbf{0.39} &  \textbf{30.15} & \textbf{0.196}  & \textbf{0.840} & \textbf{0.917}\\
\hline
\end{tabular}
\end{center}
\label{table:5}
\end{table}

\begin{table}[!htbp]
\footnotesize
\caption{Performance comparisons on the ACDC dataset.}
\vspace{2mm}
\begin{center}
\begin{tabular}{*5c}
\hline
{} &  MSE(\textit{$10^{-1}$}) & PSNR & NRMSE & SSIM\\
\hline
MFIN & 1.082 & 30.69 & 0.309  & 0.607\\ 
Slomo & 1.001 & 31.08 & 0.296  & 0.630\\ 
3D-Slomo & 0.341 & 35.27 & 0.178  & 0.845\\ 
RVLI &  0.331 &  35.66 & 0.173  & 0.860\\
Ours &  \textbf{0.081} &  \textbf{41.87} & \textbf{0.085}  & \textbf{0.953}\\
\hline
\end{tabular}
\end{center}
\label{table:6}
\end{table}


\section{Conclusion}
\subsection{Summary}
We presented a novel interpolation method for 4D dynamic medical images. Our proposed two-stage network was designed to exploit the volumetric medical images that exhibit large variations between the motion sequences. Experimental results demonstrated that our SVIN outperformed state-of-the-art temporal medical interpolation methods and natural video interpolation methods that has been extended to support volumetric images. Our ablation study further exemplified that our motion network with our adaptive multi-scale architecture was able to better represent the large functional motion compared with the state-of-the-art unsupervised medical registration methods.

\subsection{Extensions implementation}
In Section 4, we discussed our general multi-scale architecture for learning the spatial appearance volume in different scales to retain the spatial information for volume synthesis. Rather than learning a spatial transform model, in the future we will implement our architecture in other volume synthesis task. 

We leverage a regression based constrain module to explore the potential rule of functional motion. This could be extended to the other 4D volumetric task. Also, this part can be further optimized to constrain the interpolation.

Although we demonstrated our SVIN model on cardiac imaging modality, there is no restriction of our method to be applied to other dynamic images. We suggest that our method is broadly applicable to other medical, and non-medical image volume interpolation problems where the motion field can be modelled.

{\small
\bibliographystyle{ieee_fullname}
\bibliography{egbib}
}

\end{document}